\documentclass[journal]{IEEEtran}
\ifCLASSINFOpdf
\else
\fi
\usepackage[misc]{ifsym} 
\usepackage[group-separator={,}]{siunitx} 

\usepackage{amsfonts, bm, amsmath} 
\usepackage[switch]{lineno}  
\usepackage{multirow} 
\usepackage{array}
\usepackage{hhline}  
\usepackage{threeparttable}  
\usepackage{booktabs} 
\usepackage{graphicx}  
\usepackage{float} 
\usepackage{subfigure} 
\usepackage{color} 
\usepackage[colorlinks, linkcolor=blue, anchorcolor=blue, citecolor=blue]{hyperref}
\usepackage[numbers,sort&compress]{natbib}
\usepackage{tabularx}

\makeatletter  
\newcommand*\bigcdot{\mathpalette\bigcdot@{.5}}
\newcommand*\bigcdot@[2]{\mathbin{\vcenter{\hbox{\scalebox{#2}{$\m@th#1\bullet$}}}}}
\makeatother
\begin{document}
%
\title{Large-Scale Spatio-Temporal Person Re-identification: Algorithm and Benchmark}
%
%
%

\author{Xiujun~Shu$^{\dagger}$,	
    	Xiao Wang$^{\dagger}$,    	
    	Xianghao Zang,
    	Shiliang Zhang,
    	Yuanqi Chen,
        Ge~Li*,
        and Qi Tian
\thanks{$^{\dagger}$Equal Contribution. *Corresponding author.(e-mail: geli@ece.pku.edu.cn)}
\thanks{Xiujun Shu is with Peng Cheng Laboratory and Peking University, Shenzhen, China.(e-mail: shuxj@pcl.ac.cn)}
\thanks{Xiao Wang is with Peng Cheng Laboratory, Shenzhen, China. (e-mail: wangx03@pcl.ac.cn)}
\thanks{Xianghao Zang, Yuanqi Chen and Ge Li are with School of Electronic and Computer Engineering, Peking University, Shenzhen, China. (e-mail: zangxh@pku.edu.cn, cyq373@pku.edu.cn, geli@ece.pku.edu.cn)} 
\thanks{Shiliang Zhang is with School of Electronic Engineering and Computer Science, Peking University, Beijing, China. (e-mail: slzhang.jdl@pku.edu.cn)} 
\thanks{Qi Tian is with the Huawei Cloud \& AI, Huawei Technologies, China. (e-mail: tian.qi1@huawei.com)}
}

\maketitle

\begin{abstract}
	Person re-identification (re-ID) in the scenario with large spatial and temporal spans has not been fully explored. This fact partially occurs because existing benchmark datasets were mainly collected with limited spatial and temporal ranges, {\itshape e.g.,} using videos recorded in a few days by cameras in a specific region of the campus. Such limited spatial and temporal ranges make it hard to simulate the difficulties of person re-ID in real scenarios. In this work, we contribute a novel \textbf{La}rge-scale \textbf{S}patio-\textbf{T}emporal (\textbf{LaST}) person re-ID dataset, including 10,862 identities with more than 228k images. Compared with existing datasets, LaST presents more challenging and high-diversity re-ID settings and significantly larger spatial and temporal ranges. For instance, each person can appear in different cities or countries, and in various time slots from day to evening, and in different seasons from spring to winter. To our best knowledge, LaST is a novel person re-ID dataset with the largest spatio-temporal ranges. Based on LaST, we verified its challenge by conducting a comprehensive performance evaluation of 14 re-ID algorithms. We further propose an easy-to-implement baseline that works well in such challenging re-ID settings. We also verified that models pre-trained on LaST can generalize well on existing datasets with short-term and cloth-changing scenarios. We expect LaST to inspire future works toward more realistic and challenging re-ID tasks. More information about the dataset is available at \textcolor{magenta}{\url{https://github.com/shuxjweb/last.git}}.
\end{abstract}

\begin{IEEEkeywords}
	Person Re-identification, Person Retrieval, Person Recognition, Benchmark.
\end{IEEEkeywords}

%
\IEEEpeerreviewmaketitle

\section{Introduction} \label{introduction}
\IEEEPARstart{P}{erson} re-identification (re-ID), as a sub-task of image retrieval \cite{huang2020embedding, Optimal2021zhai, Deep2021zhai}, aims to match the same person in non-overlapping cameras \cite{leng2019a, 2019State, yadav2020person, ye21reidsurvey}. Existing benchmarks focus on short-term person re-ID task, {\itshape e.g.,} CUHK03 \cite{2014DeepReID}, Market1501 \cite{zheng2015scalable}, and DukeMTMC \cite{ristani2016performance} have played important roles in promoting re-ID research in recent years. However, performance on these benchmarks is becoming saturated, {\itshape e.g.,} the Rank1 accuracy exceeds 96\% on Market1501 \cite{Discriminative2019zhou, Viewpoint2020zhihui, Huang2021Multiscale}. Notwithstanding their remarkable success on those datasets, current person re-ID algorithms still exhibit limitations in real applications, where the query person may appear at different locations and wear different clothes. Therefore, a large gap between these benchmark datasets and practical applications still exists.

\begin{figure}[t]
	\centering
	\includegraphics[width=\linewidth]{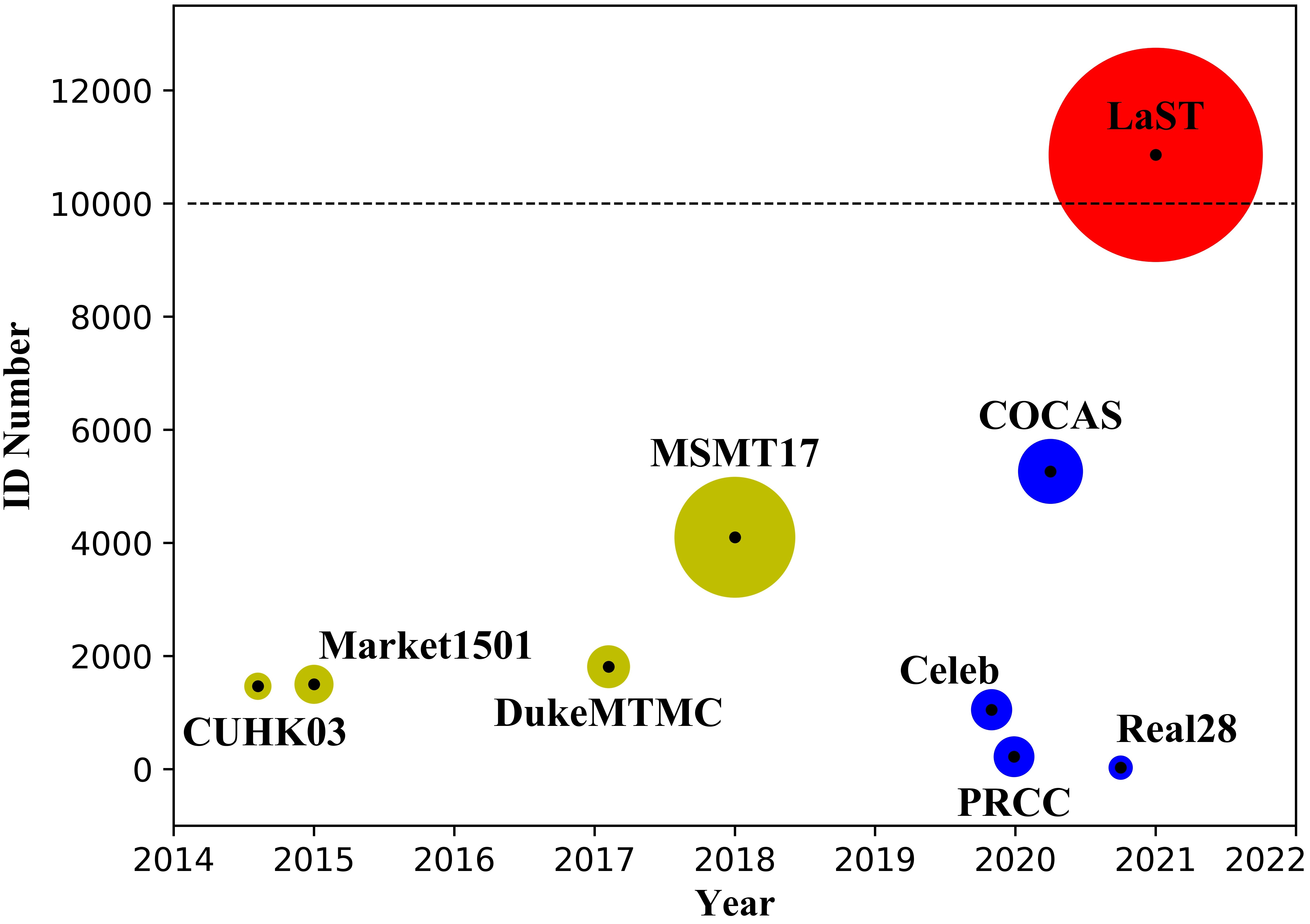} 
	\caption{\textbf{Summaries of densely annotated re-ID benchmarks.}
		The vertical axis represents the number of identities and the circle diameter is in proportion to the number of images. The yellow cycles are conventional re-ID benchmark datasets. The blue circles are cloth-changing datasets.
	}
	\label{fig:dataset}
\end{figure}

According to observation, there are two gaps between conventional benchmarks and real scenarios. 
\textbf{First}, the spatial scope of pedestrian activities is small. Most existing benchmarks were collected in a local region of the campus. The spatial scope is relatively small. This setting is quite different from the setting of real-world scenarios. For example, a suspect commonly need to be searched across cameras in several districts or the entire city.  
\textbf{Second}, the time span of pedestrian activities is short. Most of existing re-ID datasets define a short-term re-ID task, where the weather and clothes stay stable. This simplified setting also differs with the real one, where the suspects may appear at both daytime and night, and change clothes. 
The two gaps hinder the further development of person re-identification.

\begin{figure}[t]
	\centering
	\includegraphics[width=\linewidth]{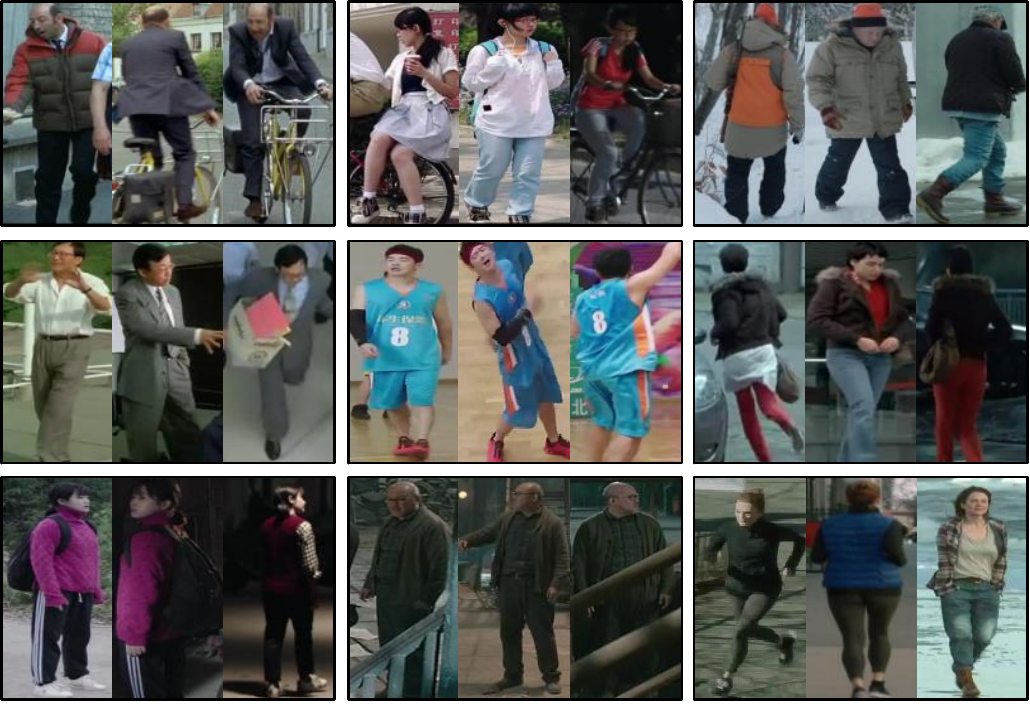} 
	\caption{\textbf{Sample images of the LaST dataset.}
		Nine persons are shown and each person has three instances.
	}
	\label{fig:instances}
\end{figure}

To address this issue, some scholars have released several cloth-changing datasets recently, {\itshape e.g.,} PRCC \cite{2021Person}, COCAS \cite{yu2020cocas} and Real28 \cite{2020When}. However, those datasets are still small-scale and cover limited spatial regions. For example, the Real28 \cite{2020When} dataset only collects images of 28 persons, the PRCC \cite{2021Person} are captured in the indoor environment, and the pedestrian images of COCAS \cite{yu2020cocas} dataset only last for four days. These limitations make them hard to simulate real-world scenarios for person re-ID. The re-ID community urgently needs a larger and more challenging dataset.

In this work, we study a more realistic person re-ID setting, which presents larger spatial and temporal ranges. We contribute a large-scale benchmark, named \textbf{LaST}, which contains 10,862 identities with 228,156 images. As shown in Fig.~\ref{fig:dataset}, LaST is the first densely annotated re-ID benchmark with more than 10k identities. Creating a large-scale person re-ID dataset covering long-range spatial and temporal spans is challenging for both video collection and data annotation. To make this procedure feasible, we explored a novel strategy by utilizing movies, which contain plenty of person images. Specifically, LaST is collected from detected persons in 2k movies. As most movies contain realistic scenes and characters, LaST is a valid benchmark for person re-ID algorithms. Moreover, compared with previous dataset construction methods, this strategy demonstrates the following advantages: 
\begin{itemize}
	\item \textbf{Privacy Protection}. Using persons detected in movies is superior to previous methods in privacy protection, which is increasingly important nowadays. 
	\item \textbf{Diversity}. As shown in Fig.~\ref{fig:instances}, scenes in movies are diverse, {\itshape e.g.,} contain both indoor, outdoor scenes. The same person can appear in different cities, counties, and wear various clothes. 
	\item \textbf{Annotations}. Annotating persons in each movie is easier than the annotation in large-scale surveillance videos. This property ensures a budget-aware dataset construction.
\end{itemize}

Although LaST comes from movies, its style is very similar to existing datasets. This fact is because eight labelers are involved to select images that have similar viewpoints to real surveillance scenes. LaST has many real challenges involved in practical scenarios. Besides the usual problems of occlusion and viewpoint, LaST has more variations in light, weather, indoor and outdoor, day and night, clothing changes. These challenges lead to significant changes in visual features of pedestrian appearances. It is difficult to retrieve the positive samples with large visual variations. This would lead to a low mean average precision (mAP). Different from current classification and metric learning methods, we proposed an efficient baseline method that directly optimizes the retrieval mAP during training. This method could retrieve more positive samples and boost the mAP to some extent. Compared with previous works \cite{vishwakarma2018deep, 2019Hierarchical, Top2020zhang, Hierarchical2021Li}, that commonly use triplet loss \cite{Wang2014Learning} or cross-entropy loss for training, this method achieves competitive performance compared with the state-of-the-art ones.

To evaluate the LaST dataset, we first test the state-of-the-art methods on LaST. Experimental results show that LaST is more challenging than existing re-ID benchmarks. Next, we conducted extensive experiments to validate the generalization performance of LaST to traditional and cloth-changing datasets. Also, we did a lot of analysis and visualization to show why we got these results. Finally, we propose a pre-training strategy based on LaST. Extensive experiments show that, pre-trained model on LaST demonstrates superior generalization ability on existing datasets, {\itshape i.e.,} both the short-term datasets and those cloth-changing ones.

\begin{figure}[t]
	\centering  
	\includegraphics[width=\linewidth]{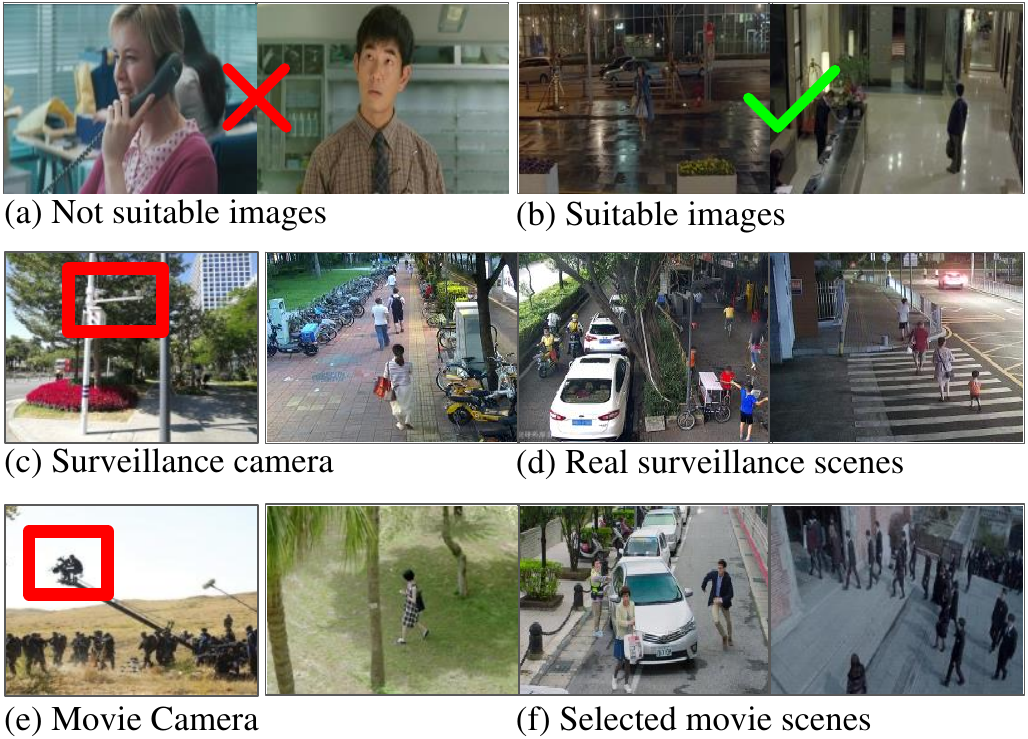}  
	\caption{\textbf{Comparison of movie scenes and surveillance scenes.}}	
	\label{fig:data_collection}
\end{figure}

In summary, our contributions are three-fold: 
\textbf{(1)} We contribute a large-scale benchmark dataset named \textbf{LaST} for person re-ID. It consists of 10,862 identities and 228,156 images in total. LaST is highly diverse capturing from a broad range of countries, person ages, scenes, weather, daytime and night. Besides, it is the first one to label clothes to date. 
\textbf{(2)} We propose an efficient baseline approach that works well on such challenging person re-ID setting. We also report the results of 14 recent person re-ID algorithms based on LaST for future works to compare.
\textbf{(3)} We conduct extensive experiments to verify the generalization ability of LaST on other re-ID datasets.

d
\begin{figure*}[t]
	\centering   
	\includegraphics[width=16cm]{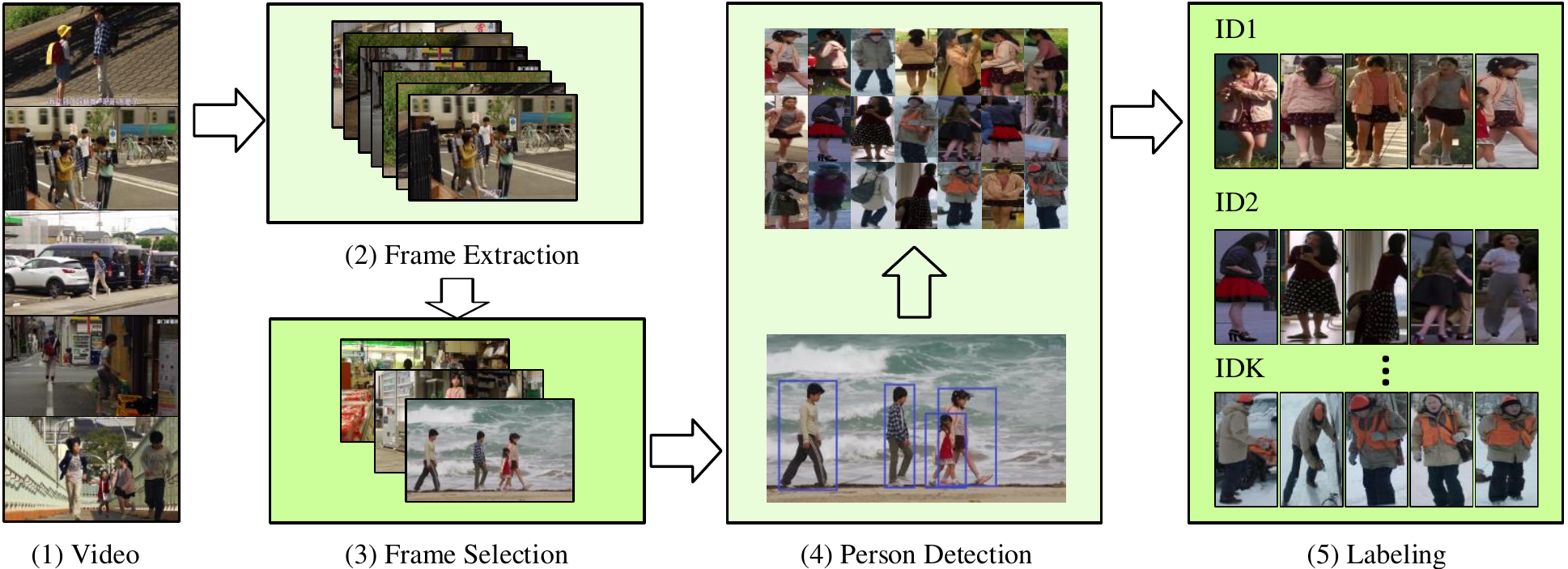} 
	\caption{\textbf{The pipeline of data building.} The second and fourth steps are processed by PLabel. The third and fifth steps require human participation. Finally, the PLabel automatically records the results of manual annotation.}	
	\label{fig:data_making}
\end{figure*}

\begin{figure*}[!t]
	\centering  
	\includegraphics[width=16cm]{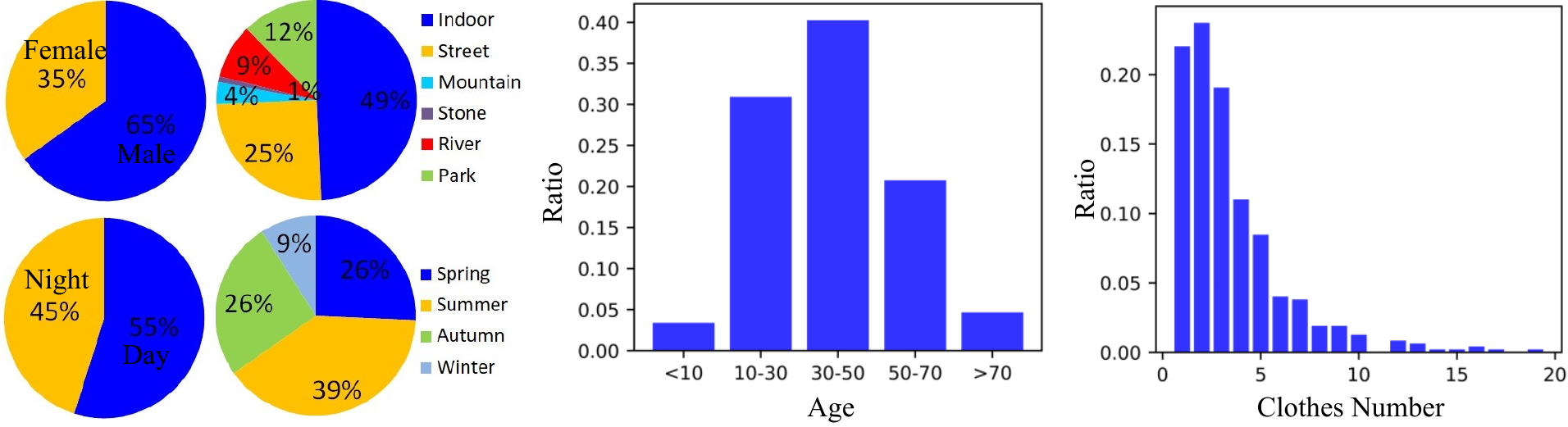} 
	\caption{\textbf{Statistical information of LaST.} All the information was counted by labelers. The ages were estimated. The stone denotes the area full of sand or stones.}	
	\label{fig:data_count}
\end{figure*}

\section{Related Work}
\subsection{Short-Term Re-ID Benchmarks}
Most re-ID benchmarks have contributed to this field in different stages. In early stages, GRID \cite{2014Person} contains 1,025 and 1,275 images captured by eight disjoint cameras in a busy underground station. CUHK01  \cite{Li2012Human} was collected on campus, which contains 971 identities and 3,884 images. The shortcoming of these datasets is that the quantity of images is too small. In later years, several larger datasets, {\itshape e.g.,} CUHK0 3\cite{2014DeepReID}, Market1501\cite{zheng2015scalable}, DukeMTMC \cite{ristani2016performance}, and MSMT17 \cite{wei2018person}, have been popularly studied in the re-ID community. Market1501 \cite{zheng2015scalable} was collected in Tsinghua University. It has 1,501 identities and 31,466 images in total. DukeMTMC \cite{wei2018person} was collected in Duke University. It contains 1,812 identities and 36,411 images. As these datasets were collected in campus, most pedestrians are college students or faculties. Besides, the performance on these datasets is becoming saturated to date. For example, the Rank1 value exceeds 96\% on Market1501 \cite{Discriminative2019zhou, Huang2021Multiscale} and reaches 91.6\% on DukeMTMC \cite{Viewpoint2020zhihui}. Besides, all these datasets focus on local space and short-term settings. They assume that the clothes of pedestrians would not change significantly. This assumption has a certain gap with real scenes and would limit the application in real-world scenarios.

\subsection{Cloth-Changing Re-ID Benchmarks}
Recently, some scholars in this field have made some efforts on the long-term re-ID \cite{huang2019beyond} and released several cloth-changing datasets. They assume that the query and the gallery of the same identity have different clothes. PRCC \cite{2021Person} includes 221 identities and 33,698 images, which capture several students in the teaching room. This dataset aims to exploit the contour sketch in the cloth-changing scenario. COCAS \cite{yu2020cocas} contains 5,266 identities and 62,384 images. It uses a cloth-template as a reference to retrieve the target person. This setting reduces the difficulty of cloth-changing person re-ID. Real28 is a small-scale benchmark consisting of 28 identities and 4,324 images. LTCC \cite{qian2020long} contains 152 identities and 17,138 images. The above datasets mainly focus on the cloth-changing issue in the time dimension. The activity scopes of pedestrians in space are limited because only a few people are recruited. Besides, the size of these datasets is still small.

\begin{figure*}[!t]
	\centering  
	\includegraphics[width=16cm]{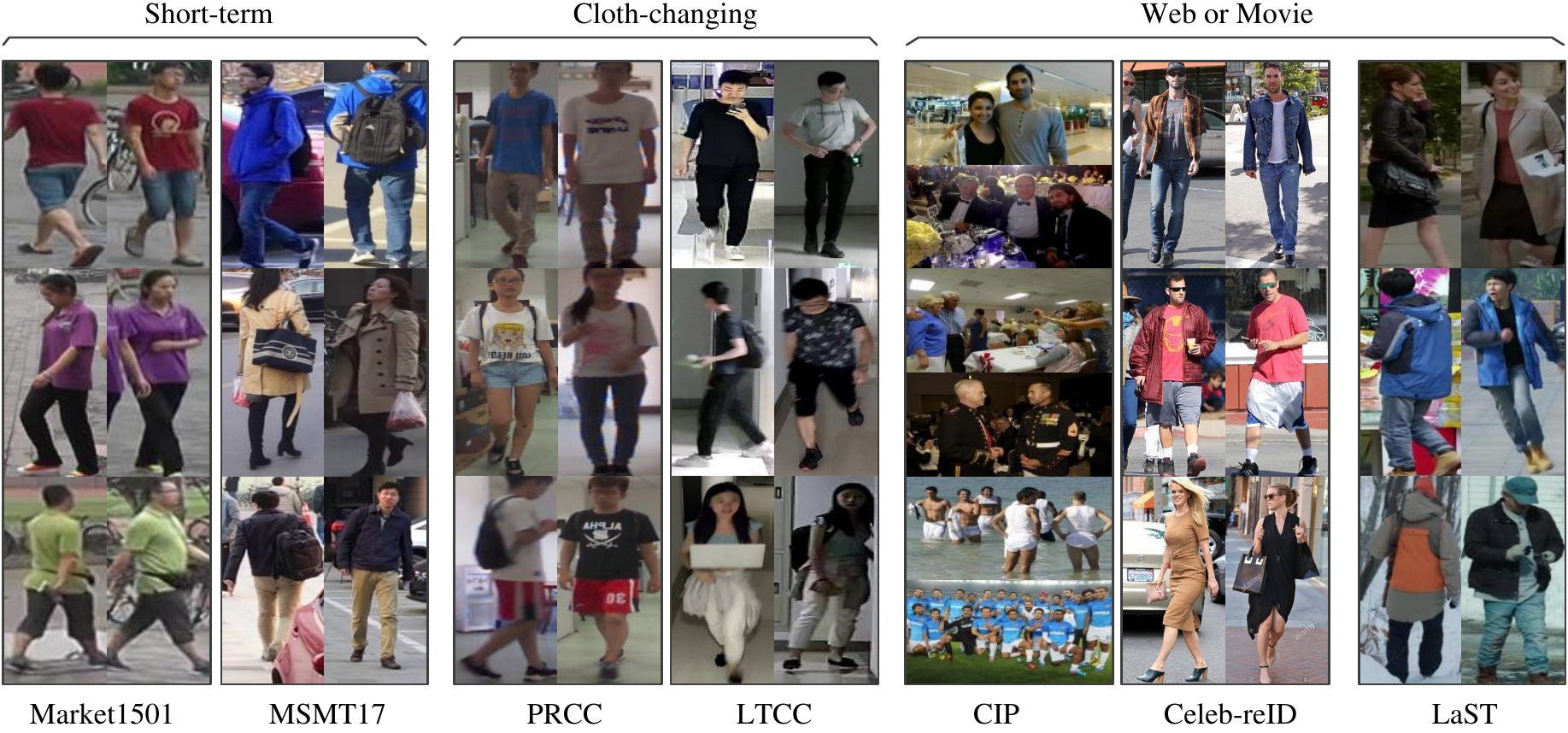}  
	\caption{\textbf{The style comparison of different benchmarks.} Market1501 and MSMT17 were collected on campus. PRCC and LTCC were collected indoors. CIP and Celeb-reID were acquired from the Internet.}	
	\label{fig:data_style}
\end{figure*}

\begin{figure*}[!t]
	\centering  
	\includegraphics[width=16cm]{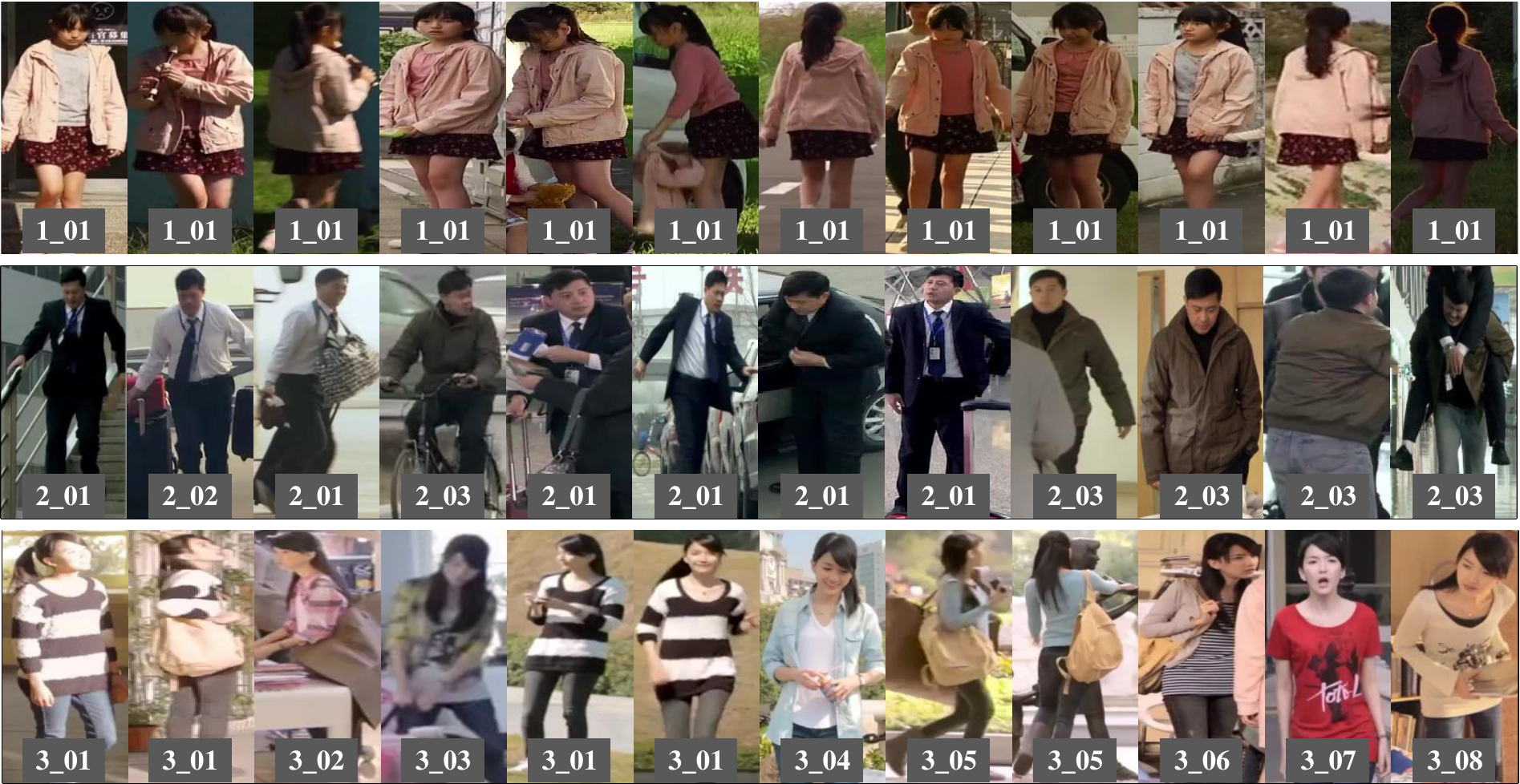}  
	\caption{\textbf{The illustration of clothing labels.} Each row denotes the same person. A\_B denotes the labels, in which A is the identity label and B is the clothing label. The little girl in the first row does not change her clothes. All the clothing labels are 01. The man in the second row has a total of 3 suits. The girl in the third row has eight clothes.}	
	\label{fig:cloth_id}
\end{figure*}

\subsection{Web or Movie-based Re-ID Benchmarks}
There are several related datasets from photo albums, short videos or movies to date. PIPA \cite{zhang2015beyond} contains 2,356 identities, which was collected from public photo albums. CIP \cite{Zhong2016Faces} contains 4,204 identities, which was collected from the Internet. CIM \cite{huang2018unifying} consists of 1,218 identities, which was collected from movies. However, the pedestrians in these datasets are not cropped images, but in the ``wild''. They were designed not for person re-ID, but for person retrieval or person search. 
MovieNet \cite{huang2020movienet} contains 3,087 identities. Its shortcoming is that many samples in MovieNet only have heads. It is not specifically designed for  person re-ID task, but for story-based long video understanding. CUHK-SYSU \cite{xiao2017joint} is used for person search, which was collected by hand-held cameras and snapshots from movies. SYSU-30k \cite{2020Weakly} was collected from Internet and TV programs. It is a large scale dataset and contains 30k identities. But its annotation is bag-level labeling, other than identity-level. The labels are not accurate enough. It is used for weakly supervised re-ID task. Celeb-reID \cite{huang2019beyond} contains 1,052 identities, which was acquired from the Internet using street snapshots of celebrities. It is the most similar dataset to ours, but still significantly different. Its shortcoming is that most pedestrians are on the frontal view. Also, the light is always bright and lacks diversity. In contrast, the images in LaST was carefully selected and labeled. The viewpoints are similar to surveillance scenes. The variations in pedestrian pose, illumination, and scenes are quite diverse.

\section{Benchmark Building and Statistics}\label{sect:hierarchy}
LaST aims to offer the community a large-scale benchmark covering long-range spatial and temporal spans. It was collected from movies and built by a self-developed semi-automatic annotation tool called PLabel, which had been open-sourced \footnote{\textcolor{magenta}{\url{https://code.ihub.org.cn/projects/4420}}}. More details about dataset building and statistics are shown as follows.

\begin{table*}[!t] 
	\centering
	\renewcommand\arraystretch{1.4}
	\caption{\textbf{Comparison of LaST with previous densely labeled re-ID benchmarks.} ``Crop'' means detected bounding box. ``Body'' means similar view angles to surveillance scenes and the whole body can be seen. ``Time'' means long-term. ``Space'' means large activity scope. ``Clothing ID'' is the label of clothes.}
		\begin{tabular}{p{2.5cm}|p{1cm}<{\centering}|p{1cm}<{\raggedleft}|p{1cm}<{\raggedleft}|p{1cm}<{\centering}|p{1cm}<{\centering}|p{1cm}<{\centering}|p{1cm}<{\centering}|p{1cm}<{\centering}|p{1.5cm}<{\centering}}
			\hline  
			\textbf{Dataset} & \textbf{Year} & \textbf{ID Num} & \textbf{Images}  & \textbf{Cameras} &\textbf{Crop} & \textbf{Body} & \textbf{Time} &  \textbf{Space}  & \textbf{Clothing ID}\\
			\hline \hline
			\textbf{VIPeR}\cite{2008Viewpoint}	
			&2007  &632    & 1,264   & 2  & $\checkmark$ &  & & \\
			
			\textbf{GRID}\cite{2014Person}	
			& 2009 &1,025  & 1,275   & 8  & $\checkmark$ &  & & \\
						
			\textbf{CUHK01}\cite{Li2012Human}	
			& 2012 &971    & 3,884   & 2  & $\checkmark$ & $\checkmark$ &  &  & \\
			
			\textbf{CUHK03}\cite{2014DeepReID}	
			& 2014 & 1,467 & 13,164  & 10 & $\checkmark$ & $\checkmark$ &  &  &\\
			
			\textbf{Market1501}\cite{zheng2015scalable}
			& 2015 & 1,501 & 31,466  & 6  & $\checkmark$ & $\checkmark$ &  &  & \\
			
			\textbf{DukeMTMC}\cite{ristani2016performance}
			& 2017 & 1,812 & 36,411  & 8  & $\checkmark$ & $\checkmark$ &  &  & \\
			
			\textbf{MSMT17}\cite{wei2018person}	
			& 2018 & 4,101 & 124,068 & 6  & $\checkmark$ & $\checkmark$ &  &  &\\
						
			\hline
			
			\textbf{PIPA}\cite{zhang2015beyond}	
			& 2015 & 2,356 & 37,107 & * & $\checkmark$ &  & $\checkmark$ & $\checkmark$ &\\
			
			\textbf{CIP}\cite{zhang2015beyond}	
			& 2016 & 4,204 & 37,916 & * & $\checkmark$ &  & $\checkmark$ & $\checkmark$ &\\
			
			\textbf{CIM}\cite{huang2018unifying}	
			& 2018 & 1,218 & 72,875 & * & $\checkmark$ &  & $\checkmark$ & $\checkmark$ &\\
			
			\textbf{Celeb-reID}\cite{huang2019beyond}	
			& 2019 & 1,052 & 34,186 & * & $\checkmark$ & $\checkmark$ & $\checkmark$ & $\checkmark$ &\\
			
			\textbf{Celebrities-reID}\cite{huang2019celebrities}	
			& 2019 & 590 & 10,842 & * & $\checkmark$ & $\checkmark$ & $\checkmark$ & $\checkmark$ &\\
			

			\hline
			
			\textbf{PRCC}\cite{2021Person}	
			& 2020 &221 & 33,698 & 3 & $\checkmark$ & $\checkmark$ &  $\checkmark$& & \\
			
			\textbf{COCAS}\cite{yu2020cocas}		
			& 2020 & 5,266 & 62,382  & 30 & $\checkmark$ & $\checkmark$ & $\checkmark$ & & \\
			 
			\textbf{Real28}\cite{2020When}	
			& 2020 &28 & 4,324 & 4 & $\checkmark$ & $\checkmark$ & $\checkmark$ &  &\\
			
			\textbf{LTCC}\cite{qian2020long}	
			& 2020 &152 & 17,138 & 12 & $\checkmark$ & $\checkmark$ & $\checkmark$ &  &\\

			\hline
			\textbf{LaST}  
			& 2021 & 10,862 & 228,156 & * & $\checkmark$ & $\checkmark$ & $\checkmark$ & $\checkmark$ & $\checkmark$ \\
			 
			\hline  
		\end{tabular}\label{table:data_compare}
\end{table*}

\subsection{Dataset Building}
Due to privacy protection, it is difficult to obtain regional or city level monitoring data in reality. Therefore, most current re-ID benchmarks were collected in local space by recruiting few people. Movies are good material for making re-ID datasets due to several reasons, {\itshape e.g.,} real pedestrians, diverse scenes, rich poses, and illumination changes. Besides, people in movies move around with a wide scope and a long-time span. 

However, most images in movies are not suitable because only the head or upper body can be seen (see Fig.~\ref{fig:data_collection}(a)). Therefore, carefully selected images that show most of the actors’ bodies and have a similar view angle to surveillance scenes were utilized (see Fig.~\ref{fig:data_collection}(b)). It is the biggest difference between LaST and other Internet-based datasets. As shown in Fig.~\ref{fig:data_collection}(c), most surveillance cameras are installed at a height of 3 to 5 meters. In movies, some cameras using high angle shot have similar heights (see Fig.~\ref{fig:data_collection}(e)). As shown in Fig.~\ref{fig:data_collection}(d) and (f), the selected movie scenes are similar to real surveillance scenes. It should also be noticed that some movies will be post-processed, which leads to changes of color and illumination to some extent. This operation can be regarded as the data augmentation. The generalization ability of LaST has been verified in the following experiments.

To ensure diversity, more than 2k movies were collected, covering more than 8 countries from Asia to Europe. Eight labelers were recruited to complete the annotation work for 2.5 months. All the annotations were done using the newly developed tool called PLabel. The pipeline of data building is shown in Fig.~\ref{fig:data_making}. First, the frames were extracted by PLabel. To reduce redundancy, the frames were then selected every several seconds. Besides, only one frame from each scene or perspective was kept. In this way, it can be considered that each frame is captured by a different camera. Next, the PLabel detected persons \cite{2019FreeAnchor} and marked the bounding boxes. Finally, each bounding box is assigned an ID number. The same person is assigned the same ID. The detected bounding boxes will be manually adjusted if the labelers observe substantial detection errors. After that, persons with less than five images were removed.

\begin{figure*}[t]
	\centering  
	\includegraphics[width=\linewidth]{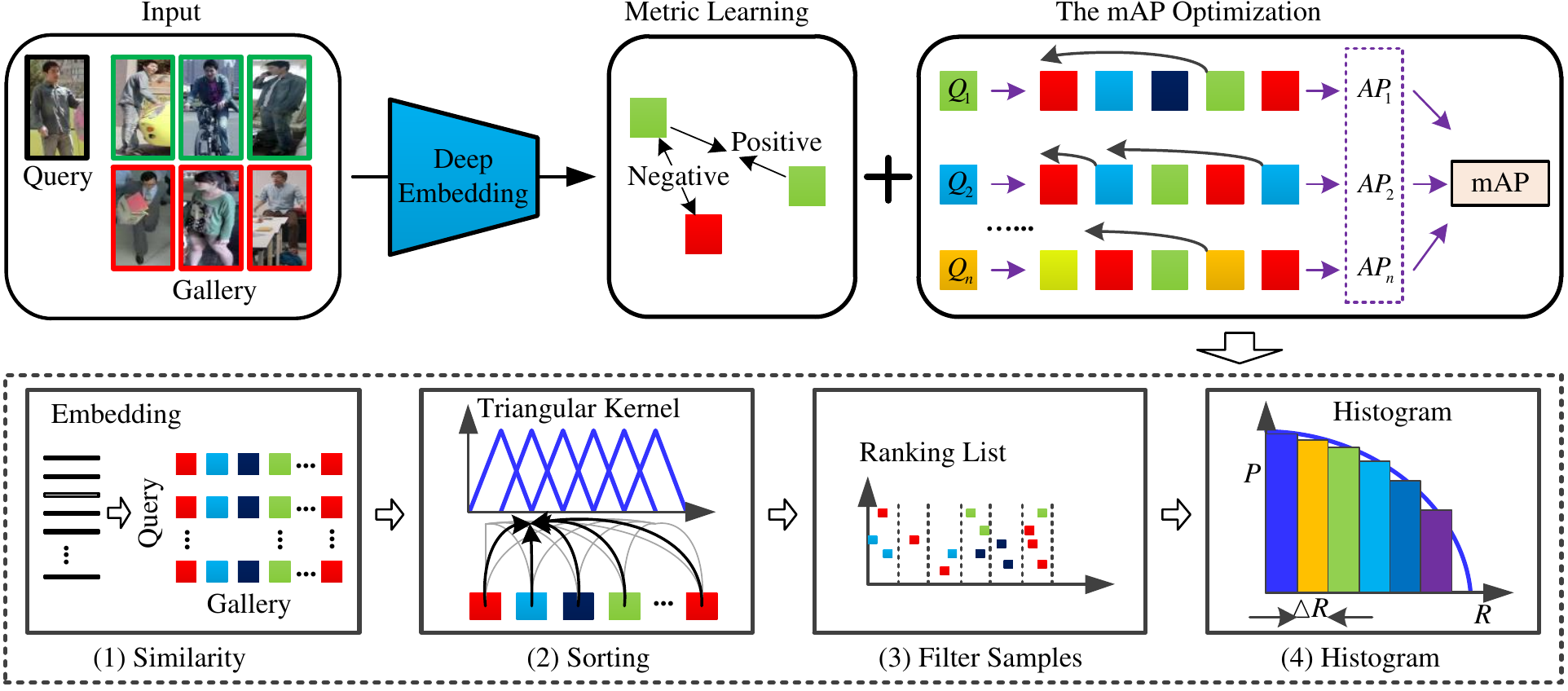}  
	\caption{\textbf{The flowchart of our approach.} 
		Metric learning utilizes triplet loss to optimize the model. This is the same with existing re-ID methods. The mAP optimization utilizes histogram approximation to directly optimize the mAP accuracy.}	
	\label{fig:framework}
\end{figure*}

\subsection{Statistical Characteristics}
Fig.~\ref{fig:data_count} gives some statistical characteristics. All the information was collected manually. Among the pedestrians, 65\% are male and 35\% are female. Many of them move among cities and even countries. The activity scenes are roughly divided into six categories: indoor, street, mountain, stone, river, and park. The stone denotes the area full of sand or stones. The river means a place where there is water, {\itshape e.g.,} river, lake, pool, beach. The large time span brings about two changes: clothes and the weather, {\itshape e.g.,} daytime and night, changing seasons. As shown in Fig.~\ref{fig:data_count}, it accounts for 55\% during the daytime and 45\% at night. The season covers from spring to winter. For pedestrians, the age span ranges from children to older people over 70 years old. 76\% of them have changed their clothes. The maximum number for each person attains to 24.

\subsection{Style comparison}
Fig.~\ref{fig:data_style} compares the style of different benchmarks. As Market1501 \cite{zheng2015scalable} and MSMT17 \cite{wei2018person} were collected on campus, most pedestrians were college students or faculties. Most of the scenes are the streets and teaching buildings inside the campus. PRCC \cite{2021Person} and LTCC \cite{qian2020long} were collected by recruiting a few students to take pictures. Most scenes are indoors and the diversity of scenes is limited. CIP \cite{Zhong2016Faces} and Celeb-reID \cite{huang2019beyond} were acquired from the Internet, {\itshape e.g.,} YouTube and short videos. The samples in CIP are not cropped, so CIP is used for the task of person retrieval or person search. Celeb-reID \cite{huang2019beyond} is from the street snap-shots of celebrities and most of the scenes are streets. LaST was collected from thousands of movies. As shown in Fig.~\ref{fig:data_style}, its style is similar to short-term and cloth-changing datasets. Since movies are long videos, pedestrians in LaST have more space and a greater time span. The scenes and clothing style in LaST are much more diverse.

\subsection{Quantitative comparison}
Table~\ref{table:data_compare} gives some comparisons with existing datasets. LaST has 10,862 identities and 228,156 images, which are both larger than those of other datasets. Compared with short-term datasets, {\itshape e.g.,} CUHK03 \cite{2014DeepReID} and Market1501 \cite{zheng2015scalable},  LaST is larger in scale and covers more complicated scenes. Compared with cloth-changing datasets, {\itshape e.g.,} PRCC \cite{2021Person} and COCAS \cite{yu2020cocas}, LaST has much more scenes. The changes of clothes and weather are still more diverse in LaST. Compared with other datasets, {\itshape e.g.,} PIPA \cite{zhang2015beyond}, CIP \cite{Zhong2016Faces}, and CIM \cite{huang2018unifying}, LaST follows the popular person re-ID setting and provides cropped person images covering more complete person body. Besides, many images can only see the heads in these datasets. Celeb-reID \cite{huang2019beyond} is similar to LaST, but most images in Celeb-reID are taken from the frontal view and are bright in the daytime. LaST is more diverse in view angles and light changes than Celeb-reID. Besides, the clothing ID is labeled in LaST for each person which is shown in Fig~\ref{fig:cloth_id}. Some persons do not change clothes, but some change clothes frequently. To the best of our knowledge, LaST is currently the first person re-ID dataset providing clothing labels.

\section{Methodology}
This work mainly studies the large spatio-temporal setting, in which pedestrians have much larger activity spans in space and in time. In a larger space, there are more pedestrians with similar appearances, causing interference to retrieval. Also, a person may appear indoor or outdoor, day or night, which would change the visual feature significantly. In the long-term span, the same person may change their clothes, leading to large appearance variation. The inter and intra-variations change much more significantly than the short-term setting. It is challenging to retrieve hard positive samples with large visual variations, which would result in a low mean average precision (mAP). Therefore,  it is necessary to improve the mAP accuracy. In this section, we present a simple but effective method that directly optimizes the mAP value during training. This is different from current classification and metric learning methods. Fig.~\ref{fig:framework} shows the flowchart of the framework. More details can be seen in the following sections.

\begin{figure}[t]
	\centering  
	\includegraphics[width=4.5cm]{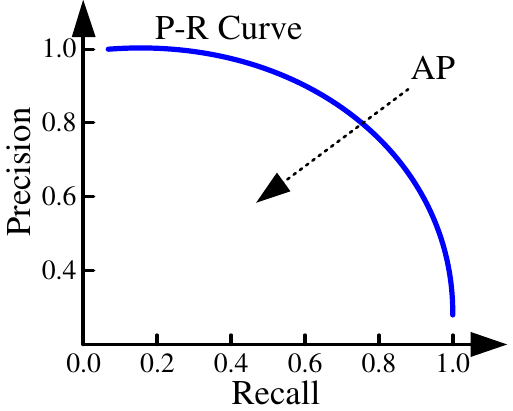}  
	\caption{\textbf{The definition of the average precision (AP).} 
		AP is defined as the area under precison-recall curve.}	
	\label{fig:pr_curve}
\end{figure} 

\subsection{Problem Formulation}
The goal of a retrieval system is to retrieve all the positive samples with the same identity in the gallery. Assume the query set is denoted as $\mathbb{Q}=\{q_i|_{i=1}^{N_\mathbb{Q}}\}$ and the gallery set is denoted as $\mathbb{G}=\{g_j|_{j=1}^{N_\mathbb{G}}\}$. Given a query image $q_i\in\mathbb{Q}$, the re-ID system ranks all instances in $\mathbb{G}$ based on the similarity or Euclidean distance to the query. The instance with larger similarity or smaller European distance ranks higher.

Most of the existing works learn a deep embedding network $f({x}|\theta)$ that embeds $x$ to a high dimensional space. To learn discriminative features, classification and metric learning are commonly used in current methods. The mean average precision (mAP) is one of the standard metrics for person re-ID task. It is the mean of the average precision (AP) of all query images. As shown in Fig.~\ref{fig:pr_curve}, AP refers to the area under the P-R curve composed of precision and recall. Therefore, the mAP can be represented as follows: 
\begin{align} 
\label{mAP}
\begin{split}
mAP&=\frac{1}{N_\mathbb{Q}}\sum_{i=1}^{N_\mathbb{Q}}AP(i), \\
&=\frac{1}{N_\mathbb{Q}}{\sum_{i=1}^{N_\mathbb{Q}}}\sum_{j=1}^{N_\mathbb{G}}\Big[P_i(j)\cdot\triangle R_i(j)\Big], 
\end{split}
\end{align}
where $N_\mathbb{Q}$ and $N_\mathbb{G}$ denote the number of the query and gallery sets. $P$ and $R$ denote the precision and recall value, respectively.

The final objective is shown as:
\begin{equation}
\label{objective}
\begin{array}{cl}
{\min } & {1 - mAP(\mathbb{Q}, \mathbb{G})}, \\ 
{\text { s.t. }} & {\mathbb{Q}=\{q_i|_{i=1}^{N_\mathbb{Q}}\}, \mathbb{G}=\{g_j|_{j=1}^{N_\mathbb{G}}\}}. 
\end{array}
\end{equation}

The above objective is a discrete optimization problem. It involves the discrete sorting operation, which is difficult for gradient-based optimization. However, we can compute an approximation of mAP by using histograms \cite{Ustinova2016Learning, He2017Hashing, brown2020smooth}. The approximate version of mAP is differentiable. 

\subsection{Learning Pipeline} 
As shown in Fig.~\ref{fig:framework}, our framework consists of metric learning\cite{Wang2014Learning} and the mAP optimization. More details are as follows.

\textbf{Metric Learning.}
The cross-entropy loss $\mathcal{L}_{i}$ and triplet loss $\mathcal{L}_{t}$ are both used to optimize the embedding network. The losses can be denoted as follows:
\begin{align}  
&\mathcal{L}_{metric}=\mathcal{L}_{i} + \mathcal{L}_{t},\\
&\mathcal{L}_{i} = \frac{1}{B}\sum_{i=1}^{B}\Big(-y_i\cdot\log p(x_i)\Big),\\
&\mathcal{L}_{t} = \frac{1}{B}\sum_{i=1}^{B}\Big(max\{m+d(x_i,x_i^p)-d(x_i,x_i^n), 0\}\Big),
\end{align} 
where $B$ is the batch size. $x_i$ and $y_i$ denote the input sample and its label, respectively. $p(x_i)$ is the predicted identity probability. $m$ is a marginal value. $d_{max}$ and $d_{min}$ denote the maximum distance of positive sample pairs and minimum distance of negative sample pairs.

\textbf{The mAP optimization.} 
Fig.~\ref{fig:framework} gives the whole pipeline of the mAP computation. To compute the AP value, the discrete sorting is needed to get the ranking list. Inspired by previous works \cite{Ustinova2016Learning, He2017Hashing, He2018Local, Jerome2019Learning}, the relaxation of sorting by can be achieved by using a group of triangular kernels. As shown in Fig.~\ref{fig:framework}, the pipeline of computing AP is divided into several steps. First, the backbone network is used to compute the embeddings of all query and gallery samples. All the features are normalized by $L_2$ norm. Then, we get the similarity matrix by computing the cosine similarity $s_{ij} \in [0,1]$ of query $q_i\in \mathbb{Q}$ and gallery $g_j\in \mathbb{G}$. The similarity matrix is composed of $\mathbb{Q}$ similarity vectors. Next, the similarity vectors for each query is input to the triangular kernels. The triangular kernel is defined as follows:
\begin{equation}  
\Delta(s_{ij}, m)=max\big(1-\frac{|s_{ij}-b_m|}{\varepsilon}, 0\big), 
\end{equation}
where $b_m$ is the $m^{th}$ bin with a range of [0, 1]. $\varepsilon$ denotes the interval between adjacent bins. Assume the number of bins is $M$, then $\varepsilon=\frac{1}{M-1}$.

For the $i^{th}$ query, its similarity vector can be denoted as $\textbf{s}=[s_{i1},s_{i2},...,s_{iN_\mathbb{G}}]$. The similarity vector $\textbf{s}$ is input to the triangular kernels. If the item $s_{ij}$ is not in the interval of [$b_m-\varepsilon, b_m+\varepsilon$], the output $\Delta(s_{ij}, m)$ would be set zero and the item $s_{ij}$ would be removed. In this way, all the similarity items in $\textbf{s}$ are sorted and a ranking list of $\textbf{s}$ can be achieved. After the above operations, all the items in $\textbf{s}$ are assigned into corresponding bins. The triangular kernels approximates the sorting process and can be realized by a convolutional operation. Therefore, the approximate sorting can be differentiable.

As shown in Fig.~\ref{fig:framework}, the histogram can be achieved after sorting. To compute the area under P-R curve, we need to get the values of the precision $P$ and recall $\triangle R$ for all bins. The precision $P$ and recall $\triangle R$ can be defined as follows:
\begin{align}  
&P_i(j) = \frac{\sum_{m=1}^{j}\sum_{n=1}^{N_\mathbb{G}}\Big(\Delta(s_{in}, j)\cdot \bar{y}_{in}\Big)}{\sum_{m=1}^{j}\sum_{n=1}^{N_\mathbb{G}}\Big(\Delta(s_{in}, j)\Big)},\\
&\triangle R_i(j) =  \frac{\sum_{n=1}^{N_\mathbb{G}}\Big(\Delta(s_{in}, j)\cdot \bar{y}_{in}\Big)}{\sum_{n=1}^{N_\mathbb{G}}\Big(\bar{y}_{in}\Big)}, 
\end{align} 
where $\triangle R_i(j)=R_i(j)-R_i(j-1)$. $s_{in}$ is the similarity between $q_i$ and $g_n$. $\bar{y}_{in}\in\{0, 1\}$ indicates whether belongs to the same identity. 

Finally, the mAP loss can be denoted as:
\begin{align} 
\label{map_loss}
\begin{split}
\mathcal{L}_{mAP}&=1 - mAP(\mathbb{Q}, \mathbb{G}), \\
&=1 - \frac{1}{N_\mathbb{Q}}{\sum_{i=1}^{N_\mathbb{Q}}}\sum_{j=1}^{N_\mathbb{G}}\Big[P_i(j)\cdot\triangle R_i(j)\Big], 
\end{split}
\end{align}

In summary, the overall loss functions can be formulated as follows:
\begin{equation} 
\mathcal{L} = \mathcal{L}_{metric} + \mathcal{L}_{mAP}.
\end{equation}

\section{Experiments} 
\subsection{Implementation Details}  

\subsubsection{Datasets}  
Three types of datasets are used in the following experiments: short-term datasets (Market1501 \cite{zheng2015scalable}, DukeMTMC \cite{ristani2016performance}, and MSMT17 \cite{wei2018person}), cloth-changing datasets (PRCC \cite{2021Person}), and spatio-temporal datasets (Celeb-reID \cite{huang2019beyond}, Celebrities-ReID \cite{huang2019celebrities}, LaST). 

Market1501 \cite{zheng2015scalable} is a widely-used person re-ID dataset. It contains 32,668 images of 1,501 identities captured by 6 cameras. Among them, 12,185 images of 751 identities are used for training. 15,913 images of 750 identities are used for testing gallery and 3,368 images are testing query. DukeMTMC-reID \cite{ristani2016performance} is captured by 8 cameras, which contains 36,411 images and 1,812 identities. MSMT \cite{wei2018person} is a large-scale dataset, which contains 124,068 images and 4,101 identities. PRCC \cite{2021Person} is a cloth-changing dataset. It contains 221 identities and 33,698 images. Celeb-reID was aquired from the Internet using street snap-shots of celebrities. It contains 1,052 identities and 34,186 images. Celebrities-ReID \cite{huang2019celebrities} contains 590 persons with 10,842 images, in which the clothes of the same person are different.

LaST contains 10,862 identities and  228,156 images. As shown in Table~\ref{table:train_eval_test}, it is split into a training set with 70,923 images, an evaluation set with 20,584 images, and a test set with 133,214 images. The identities are 5,000, 56 and 5,803, respectively.

\begin{table}[t]
	\centering
	\renewcommand\arraystretch{1.4}
	\caption{\textbf{Statistics of train/validation/test sets on LaST.}} 
	\begin{tabular}{p{1cm}|p{1cm}<{\raggedleft}|p{0.9cm}<{\raggedleft}p{0.9cm}<{\raggedleft}|p{0.9cm}<{\raggedleft}p{0.9cm}<{\raggedleft}}
		\hline
		\multirow{2}{*}{\textbf{Quantity}} & \multicolumn{1}{r|}{\textbf{Train}} & \multicolumn{2}{r|}{\textbf{Validation }} & \multicolumn{2}{r}{\textbf{Test}} \\ 
		\cline{3-6} 
		&	& Query   & Gallery   & Query   & Gallery  \\ 
		\hline 
		Identities     
		& \num{5000}  & \num{56}  & \num{56}  & \num{5805}  & \num{5806}  \\
		\hline
		Images  	       
		& \num{71248}  & \num{100}  & \num{21279}  & \num{10176}  & \num{125353} \\ 
		\hline
	\end{tabular}\label{table:train_eval_test}
\end{table}

\begin{table}[!t]
	\centering
	\renewcommand\arraystretch{1.4}
	\caption{\textbf{Re-ID results on the proposed LaST dataset}.}
	\begin{tabular}{p{2cm}|p{0.65cm}<{\centering}|p{0.55cm}<{\centering}p{0.5cm}<{\centering}p{0.5cm}<{\centering}p{0.5cm}<{\centering}p{0.5cm}<{\centering}}
		\hline 
		\textbf{Methods} &\textbf{Year} &\textbf{R1}     & \textbf{R5}   & \textbf{R10} & \textbf{R20}  & \textbf{mAP} \\ 
		\hline
		PCB\cite{sun2018beyond} &2018&50.6 &68.0 &73.9 &78.9 &15.2 \\
		
		MGN\cite{wang2018learning} &2018&41.0 &63.0 &76.0 &78.0 &17.6 \\ 
		
		SFT\cite{2019Spectral}  &2019&61.2 &75.3 &79.9 &83.5 &19.3 \\ 
		
		MHN\cite{2019Mixed}  &2019&53.7 &70.5 &76.1 &81.0 &15.4 \\ 
		
		OSNet\cite{2019Omni}   &2019&64.3 &78.9 &82.6 &86.0 &21.0 \\ 
		
		ABD-Net\cite{Chen2019ABD} &2019&48.5 &67.6 &74.4 &80.0 &16.1 \\ 
		
		BDB\cite{Dai2019Batch} &2019&62.1 &77.8 &82.4 &86.5 &19.8 \\ 
		
		PyrNet\cite{2019Aggregating}  &2019&56.4 &73.2 &78.3 &82.7 &17.2 \\ 
		
		BoT\cite{luo2019bag}  &2019&67.1 &81.0 &84.6 &87.9 &23.6 \\ 
		
		HPM\cite{fu2019horizontal} &2019&64.0 &80.0 &87.0 &89.0 &26.8 \\ 
		
		Top-DB-Net\cite{Quispe2020top} &2020&69.4 &82.8 &86.3 &89.3 &25.0 \\ 
		
		HOReID\cite{Wang2020High} &2020&68.3 &82.3 &86.2 &89.2 &25.5 \\ 
		
		CtF\cite{Wang2020Faster} &2020&70.0 &83.3 &86.7 &89.5 &26.5 \\ 
		
		QAConv\cite{liao2019interpretable} &2020 &64.6 &82.4 &79.3 &83.5 &22.4 \\

		\hline
		Ours     &2021& \textbf{71.0} &\textbf{84.3} &\textbf{87.7} &\textbf{90.5} &\textbf{28.0}\\
		
		\hline
	\end{tabular}\label{table:compare_sota}
\end{table}

\subsubsection{Experimental Setup}
ResNet50 [51] serves as the backbone network, which is initialized with ImageNet \cite{russakovsky2015imagenet} pre-trained model. The input images are resized to 256$\times$128 and augmented by random horizontal flip and random erasing. The dimension of the extracted features is 2048. The SGD optimizer is used in these experiments. The initial learning rate is $3.5\times10^{-3}$. The mini-batch size is 64, which contains 16 identities and 4 images for each identity.

\subsubsection{Evaluated Metrics}
The cumulative matching characteristic (CMC) curve and mean average precision (mAP) are utilized as evaluation metrics. CMC curve is a precision curve that provides recognition precision for each rank. mAP is the average precision value across all queries, which is much more effective than CMC when multiple ground truth exist in the gallery.

%
%
%
%
%
%
%
%

\begin{table}[!t]
	\centering
	\renewcommand\arraystretch{1.4}
	\caption{\textbf{Evaluation of proposed method on cloth-changing datasets.}}
	\begin{tabular}{p{1.7cm}|p{0.45cm}<{\centering}p{0.45cm}<{\centering}|p{0.45cm}<{\centering}p{0.45cm}<{\centering}|p{0.65cm}<{\centering}p{0.65cm}<{\centering}}
		\hline
		\multirow{2}{*}{\textbf{Methods}} &  \multicolumn{2}{c|}{\textbf{PRCC}} & \multicolumn{2}{c|}{\textbf{Celeb-reID}} & \multicolumn{2}{c}{\textbf{Celebrities-reID}} \\ 
		
		\cline{2-7} 
		& \textbf{R1}  & \textbf{mAP}  & \textbf{R1}  & \textbf{mAP}
		& \textbf{R1}  & \textbf{mAP} \\ 
		\hline
		HACNN\cite{2018Harmonious}   
		&21.8 &- &47.6 &9.5 &16.2 &11.5\\ 
		
		PCB\cite{sun2018beyond}   
		&22.9 &- &37.1 &8.2 &16.7 &9.0\\ 
		
		SPT\cite{2021Person}   
		&34.4 &- &- &- &- &-\\ 
		
		ReIDCaps\cite{huang2019beyond}   
		&- &- &51.2 &9.8 &- &-\\ 
		
		2SF-BPart\cite{huang2019celebrities}   
		&- &- &- &- &26.8 &14.0\\ 
		
		\hline
		Baseline\cite{luo2019bag}   
		&50.7 &49.8 &52.3 &9.2 &23.9 &13.1\\ 
		
		Ours  
		&\footnotesize{\textbf{57.5}} &\footnotesize{\textbf{54.7}} &\footnotesize{\textbf{54.4}} &\footnotesize{\textbf{11.8}} 
		&\footnotesize{\textbf{29.0}} &\footnotesize{\textbf{16.3}}\\ 
		
		\hline
	\end{tabular}\label{table:eval_prcc_celeb}
\end{table}

\begin{table}[!t] 
	\centering 
	\renewcommand\arraystretch{1.4} 
	\caption{\textbf{Evaluation of proposed method on short-term datasets.}} 
	\begin{tabular}{p{2.3cm}|p{0.9cm}<{\centering}p{0.9cm}<{\centering}|p{0.9cm}<{\centering}p{0.9cm}<{\centering}}
		\hline
		\multirow{2}{*}{\textbf{Methods}} &  \multicolumn{2}{c|}{\textbf{Market1501}} & \multicolumn{2}{c}{\textbf{DukeMTMC}} \\ 
		
		\cline{2-5} 
		& \textbf{R1}  & \textbf{mAP}  
		& \textbf{R1}  & \textbf{mAP} \\ 
		\hline		 
		Baseline   
		&92.2 &81.5 &84.8 &72.0 \\ 
		
		Ours  
		&93.5 &82.0 &85.7 &72.5 \\ 
		
		\hline
	\end{tabular}\label{table:eval_short_term} 
\end{table}

\subsection{Evaluation of Proposed Method}

\subsubsection{Evaluation on LaST} 
First, 14 representative algorithms are evaluated with the LaST. These algorithms achieve superior performance on conventional datasets and their codes have been released on Github. To facilitate future comparison, we report Rank1, Rank5, Rank10, Rank20, and mAP accuracies. Table~\ref{table:compare_sota} shows that the performances of existing methods are quite poor on LaST. For example, CtF \cite{Wang2020Faster} achieves 70.0\% on Rank1 accuracy and 26.5\% on mAP, which is the best among current methods. However, the mAP values of all algorithms are lower than 30\%. This result is due to the large visual variances induced by large spatial and temporal spans. For example, it is difficult to retrieve pedestrians when clothes have been changed. The experimental results demonstrate that LaST is a challenging benchmark so far. As shown in Table~\ref{table:compare_sota}, our method achieves 71.0\% on Rank1 accuracy, which is close to CtF. But our method achieves 28.0\% on mAP, which surpasses most existing methods. This benefits from the mAP optimization, which directly optimizes the mAP value during training.

\begin{figure}[!t]
	\centering  
	\includegraphics[width=\linewidth]{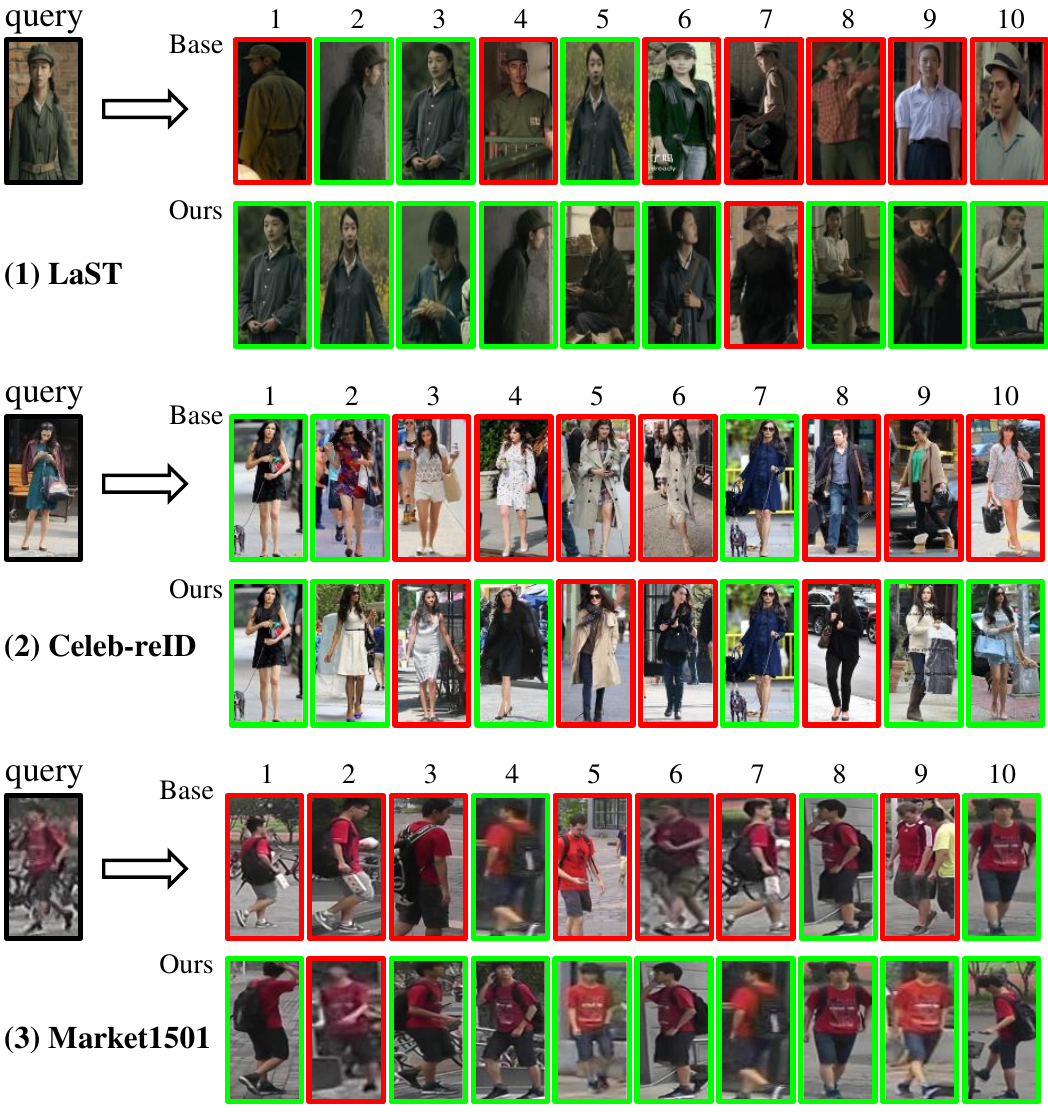} 
	\caption{\textbf{The top-10 retrieved results.} The red boxes denote negative samples, and the green boxes denote positive samples.}	
	\label{fig:rank_all}
\end{figure}

\begin{figure}[!t]
	\centering  
	\includegraphics[width=\linewidth]{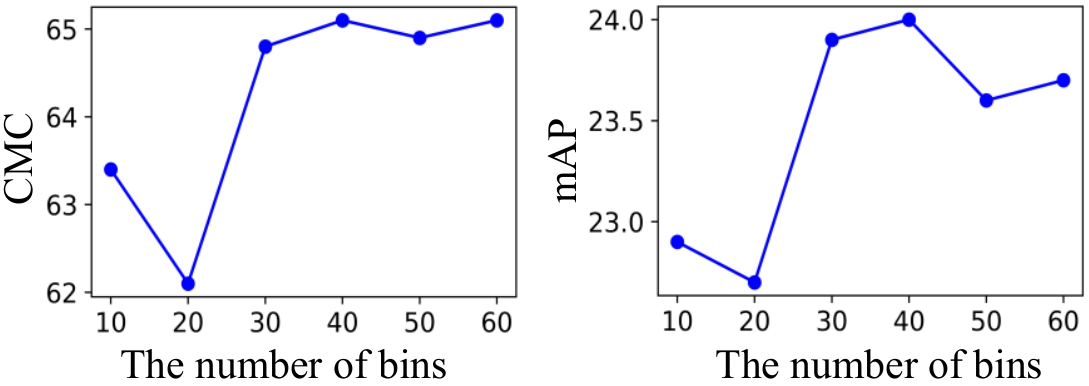} 
	\caption{\textbf{The effect of histogram bins on final performance.} }	
	\label{fig:bin_number}
\end{figure}

\subsubsection{Evaluation on Cloth-Changing Datasets}  
To verify the generalization ability of our method, experiments are further conducted on PRCC, Celeb-reID and Celebrities-reID, respectively. PRCC \cite{2021Person} is a cloth-changing dataset. Celeb-reID \cite{huang2019beyond} and Celebrities-reID \cite{huang2019celebrities} were collected from celebrity street shots. The clothes of the same identity in Celebrities-reID are different. These experiments utilize BoT \cite{luo2019bag} as the baseline and it is compared with the proposed method. As shown in Table~\ref{table:eval_prcc_celeb}, the performance of the proposed method is much better than other methods. Compared with the baseline, ours boosts the mAP accuracy by 15.5\%, 2.6\% and 3.2\% on three datasets, respectively. The above experiments demonstrate the effectiveness of the proposed method.

\subsubsection{Evaluation on Short-Term Datasets} 
Experiments were further conducted on traditional short-term datasets. As shown in Table~\ref{table:eval_short_term}, the mAP optimization has boosted the performance to some extent on Market1501 and DukeMTMC. However, the performance improvement on traditional datasets is not as significant as that on the cloth-changing datasets. This is caused by different characteristics between short-term datasets and cloth-changing datasets. The visual variations exist commonly in cloth-changing datasets, but they do not in short-term datasets. For short-term datasets, metric learning has been able to learn discriminative cues well and the performance is high enough. 

\subsection{Ranking Visualization} 
To further analyze the above results, the top-10 retrieved results on three datasets are shown in Fig.~\ref{fig:rank_all}. For LaST, the query girl is wearing a green coat. The baseline method can retrieve some positive samples with similar appearances. However, it cannot retrieve the positive samples with large visual variations, e.g., different scenes (the top-5 and top-6 in the second row) and different clothes (the top-8 and top-10 in the second row). For Celeb-reID, the proposed method can retrieve more hard positive samples, e.g., the top-2, top-3, and top-9. These samples are cloth-changing. For Market1501, the query boy was carrying a black schoolbag, wearing shorts and a dark red T-shirt. For the baseline method, most of the retrieved samples have a similar degree of brightness. The top-5 image is brighter than others but is wrong. With the help of mAP optimization, more brighter positive samples are retrieved, e.g., top-5 and top-9. The color difference in clothes is caused by the change in light.

In this work, optimizing mAP is not designed specifically for the problem of changing clothes, which is a difficult problem that needs to be further studied. We aim to provide a simple but effectiveness baseline that work on visual variations. The above experiments have demonstrated its effectiveness.

%
%

\begin{table}[t]
	\centering
	\renewcommand\arraystretch{1.4}
	\caption{\textbf{Ablation studies of our method on LaST}. $\mathcal{L}_{metric}$ denotes metric learning. $\mathcal{L}_{mAP}$ denotes the mAP optimization.} 
	\begin{tabular}{p{1cm}<{\centering}|p{1cm}<{\centering}|p{0.6cm}<{\centering}|p{0.6cm}<{\centering}|p{0.6cm}<{\centering}|p{0.6cm}<{\centering}|p{0.6cm}<{\centering}}
		\hline
		$\mathcal{L}_{metric}$ 	& $\mathcal{L}_{mAP}$   & \textbf{R1}   	& \textbf{R5}   	& \textbf{R10}   & \textbf{R20}   & \textbf{mAP} \\ 
		\hline 
		$\checkmark$    &		
		& 67.1  & 81.0  & 84.6  & 87.9  & 23.6 \\
		
		& $\checkmark$		
		& 70.4  & 83.1  & 87.6  & 90.0  & 27.5 \\
		
		$\checkmark$    & $\checkmark$  
		& 71.0  & 84.3  & 87.7  & 90.5  & 28.0 \\
		\hline
	\end{tabular}\label{table:ablation}
\end{table}

\subsection{Ablation Study}

\subsubsection{Effectiveness of the mAP Optimization}
Table~\ref{table:ablation} evaluates the metric learning and the mAP optimization, respectively. In the long-term setting, the appearance of the same person would change significantly. It is quite challenging to retrieve a person with large appearance variation. Therefore, the mAP value would be very poor. In this work, we focus on improving the mAP value. As shown in Table~\ref{table:ablation}, the mAP optimization achieves 70.4\% on Rank1 accuracy and 27.5\% on mAP. The combination of metric learning and mAP optimization achieves the best performance. In other words, the mAP optimization has indeed promoted the final performance to some extent.

\subsubsection{The Effect of Histogram Bins}
The histogram is used to calculate the mAP value. To obtain the histogram, the interval [-1,1] is first split into $M$ bins. Then the AP value is computed in each bin. As this method is an approximation of mAP, the number of bins will affect the accuracy of mAP calculation, and then affect the final performance. Fig.~\ref{fig:bin_number} shows the results of the experiments conducted on the LaST. It shows us that the Rank1 and mAP accuracies change with the number of bins. Since they achieve the best when the bin number is 40, the bin number is set 40 in our experiments.

\subsection{Generalization of LaST on Cloth-changing Setting}

\subsubsection{Direct Transfer Evaluation}
To evaluate the generalization ability of LaST, direct transfer experiments were conducted. For example, the model is trained on Market1501 and tested on PRCC dataset. In the following experiments, BoT\cite{luo2019bag} is used as the baseline method. As shown in Table~\ref{table:direct_transfer}, five large-scale datasets are used as the training sets. The two cloth-changing datasets, PRCC \cite{2021Person} and Celeb-reID \cite{huang2019beyond}, are used as the test sets. Although ImageNet has millions of images, its direct transfer performance is far inferior to other pedestrian-based datasets. The reason is that most samples in ImageNet are not persons. Learning on pedestrian datasets is more helpful in learning discriminative human cues. Table~\ref{table:direct_transfer} shows LaST achieves the best performance. For example, it achieves the Rank1 accuracy of 39.3\% on PRCC, 13.1\% higher than MSMT17. This benefits from the larger diversity in LaST, leading to a better generalization performance.

%
%
%
%
%
%
%
%

\begin{table}[!t] 
	\centering
	\renewcommand\arraystretch{1.4}
	\caption{\textbf{Performance comparison of direct transfer.}
		The model was trained and tested on different datasets.}
	\begin{tabular}{p{2.3cm}|p{0.9cm}<{\centering}p{0.9cm}<{\centering}|p{0.9cm}<{\centering}p{0.9cm}<{\centering}}
		\hline
		\multirow{2}{*}{\textbf{Training Set}} &  \multicolumn{2}{c|}{\textbf{PRCC}} & \multicolumn{2}{c}{\textbf{Celeb-reID}} \\ 
		
		\cline{2-5} 
		& \textbf{R1}  & \textbf{mAP}  
		& \textbf{R1}  & \textbf{mAP} \\ 
		\hline
		ImageNet\cite{russakovsky2015imagenet}  
		&24.7 &13.5 &28.7 &3.0 \\ 
		
		Market1501\cite{ zheng2015scalable} 
		&29.0 &24.3 &36.7 &3.7 \\ 
		
		DukeMTMC\cite{zheng2017unlabeled}  
		&28.3 &24.1 &40.9 &4.6 \\
		
		MSMT17\cite{wei2018person}  
		&26.2 &24.6 &43.4 &5.0 \\
		
		LaST  
		&\footnotesize{\textbf{39.3}} &\footnotesize{\textbf{32.6}} &\footnotesize{\textbf{47.0}} &\footnotesize{\textbf{7.0}} \\

		\hline
	\end{tabular}\label{table:direct_transfer}
\end{table}

%
%
%
%
%
%
%
%

\begin{table}[t]
	\centering
	\renewcommand\arraystretch{1.4}
	\caption{\textbf{Performance comparison of domain adaptation.}
		The model was trained and tested on PRCC and Celeb-reID, respectively.}
	\begin{tabular}{p{2.3cm}|p{0.9cm}<{\centering}p{0.9cm}<{\centering}|p{0.9cm}<{\centering}p{0.9cm}<{\centering}}
		\hline
		\multirow{2}{*}{\textbf{Pre-Training}} &  \multicolumn{2}{c|}{\textbf{PRCC}} & \multicolumn{2}{c}{\textbf{Celeb-reID}} \\ 
		
		\cline{2-5} 
		& \textbf{R1}  & \textbf{mAP}  
		& \textbf{R1}  & \textbf{mAP} \\ 
		\hline
		ImageNet\cite{russakovsky2015imagenet}  
		&43.1 &41.3 &49.2 &8.7 \\ 
		
		Market1501\cite{ zheng2015scalable} 
		&44.3 &43.1 &49.3 &8.7 \\ 
		
		DukeMTMC\cite{zheng2017unlabeled}  
		&43.9 &44.2 &49.8 &8.9 \\
		
		MSMT17\cite{wei2018person}  
		&43.7 &44.1 &51.0 &9.0 \\
		
		LaST  
		&\footnotesize{\textbf{54.4}} &\footnotesize{\textbf{54.3}} &\footnotesize{\textbf{56.1}} &\footnotesize{\textbf{11.7}} \\

		\hline
	\end{tabular}\label{table:transfer_sl}
\end{table}

%
%
%
%
%
%
%

\begin{table}[t]
	\centering
	\renewcommand\arraystretch{1.4}
	\caption{\textbf{Performance comparison on short-term datasets.}
		The model was trained and tested on Market1501 and DukeMTMC, respectively.}
		\begin{tabular}{p{2.3cm}|p{0.9cm}<{\centering}p{0.9cm}<{\centering}|p{0.9cm}<{\centering}p{0.9cm}<{\centering}}
			\hline
			\multirow{2}{*}{\textbf{Pre-Training}} &  \multicolumn{2}{c|}{\textbf{Market1501}} & \multicolumn{2}{c}{\textbf{DukeMTMC}} \\ 
			
			\cline{2-5} 
			& \textbf{R1}  & \textbf{mAP}  
			& \textbf{R1}  & \textbf{mAP} \\ 
			\hline
			ImageNet\cite{russakovsky2015imagenet}  
			&89.0 &71.1 &78.7 &62.8 \\ 
			
			MSMT17\cite{wei2018person}  
			&92.5 &79.2 &84.4 &70.7 \\
			
			\hline
			LaST  
			&91.4 &79.1 &82.5 &69.0 \\
			
			LaST\_Cloth  
			&\footnotesize{\textbf{93.1}} &\footnotesize{\textbf{81.7}} &\footnotesize{\textbf{84.5}} &\footnotesize{\textbf{71.7}} \\

			\hline
	\end{tabular}\label{table:transfer_short_term}
\end{table}

%
%
%
%
%
%

\begin{table}[!t]
	\centering
	\renewcommand\arraystretch{1.4}
	\caption{\textbf{Performance comparison of identity combination.}
		The training data contains the training set and target set. The model was tested on the target set, {\itshape i.e.,} Market1501 and DukeMTMC.}
		\begin{tabular}{p{2.3cm}|p{0.9cm}<{\centering}p{0.9cm}<{\centering}|p{0.9cm}<{\centering}p{0.9cm}<{\centering}}
			\hline
			\multirow{2}{*}{\textbf{Training Set}} &  \multicolumn{2}{c|}{\textbf{Market1501}} & \multicolumn{2}{c}{\textbf{DukeMTMC}} \\ 
			
			\cline{2-5} 
			& \textbf{R1}  & \textbf{mAP}  
			& \textbf{R1}  & \textbf{mAP} \\ 
			\hline
			
			+MSMT17\cite{wei2018person}  
			&91.7 &77.4 &83.4 &69.7 \\
			
			+LaST\_Cloth  
			&\footnotesize{\textbf{92.3}} &\footnotesize{\textbf{80.0}} &\footnotesize{\textbf{84.3}} &\footnotesize{\textbf{70.2}} \\

			\hline
	\end{tabular}\label{table:transfer_comb}
\end{table}

\begin{figure*}[t]
	\centering  
	\includegraphics[width=\linewidth]{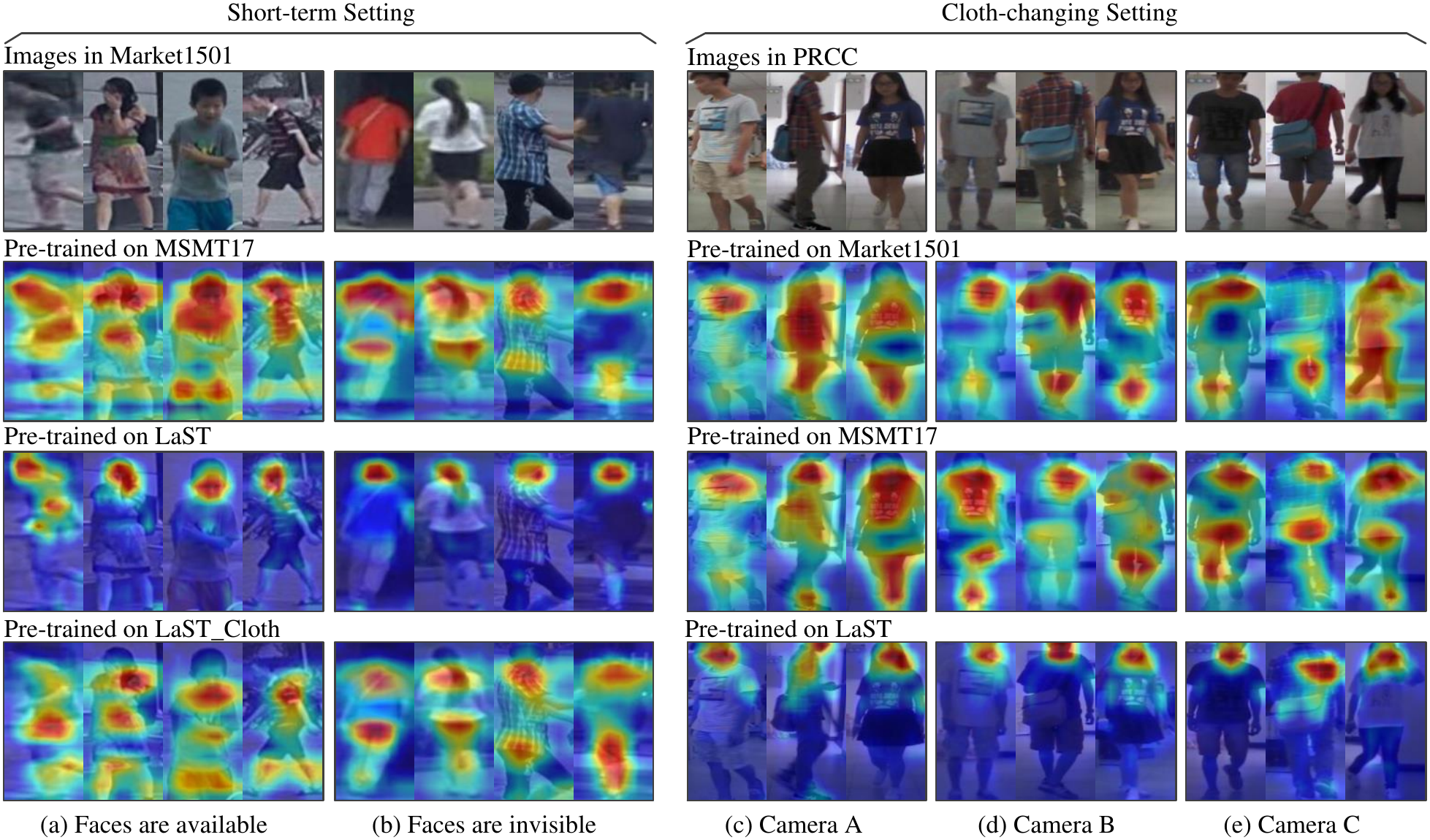} 
	\caption{\textbf{The visualization of heat maps calculated by different pre-trained models.} For example, ``Pre-trained on MSMT17'' means that the model is first trained on MSMT17. Then the trained model is used to visualize the images in Market1501.}	
	\label{fig:heat_map}
\end{figure*}

\subsubsection{Domain Adaptation}
When a few labeled samples on target domains are available, domain adaptation can be simply done by supervised fine-tuning on the pre-trained model. For example, the model is first initialized with the pre-trained parameters on Market1501, then it is fine-tuned on PRCC dataset. As shown in Table~\ref{table:transfer_sl}, five datasets are used as the source datasets. The pre-trained models on the five datasets are utilized to initialize the backbone network. Then the network is trained and tested on the target domain. We can see that the performance of all datasets has been boosted to some extent. This is benefited from the source datasets. LaST achieves 54.4\% and 56.1\% on Rank1 accuracy, respectively. It surpasses all other datasets on two target domains. The above experiments fully demonstrate the superior generalization ability of LaST on cloth-changing scenarios.

\begin{figure*}[t]
	\centering  
	\includegraphics[width=\linewidth]{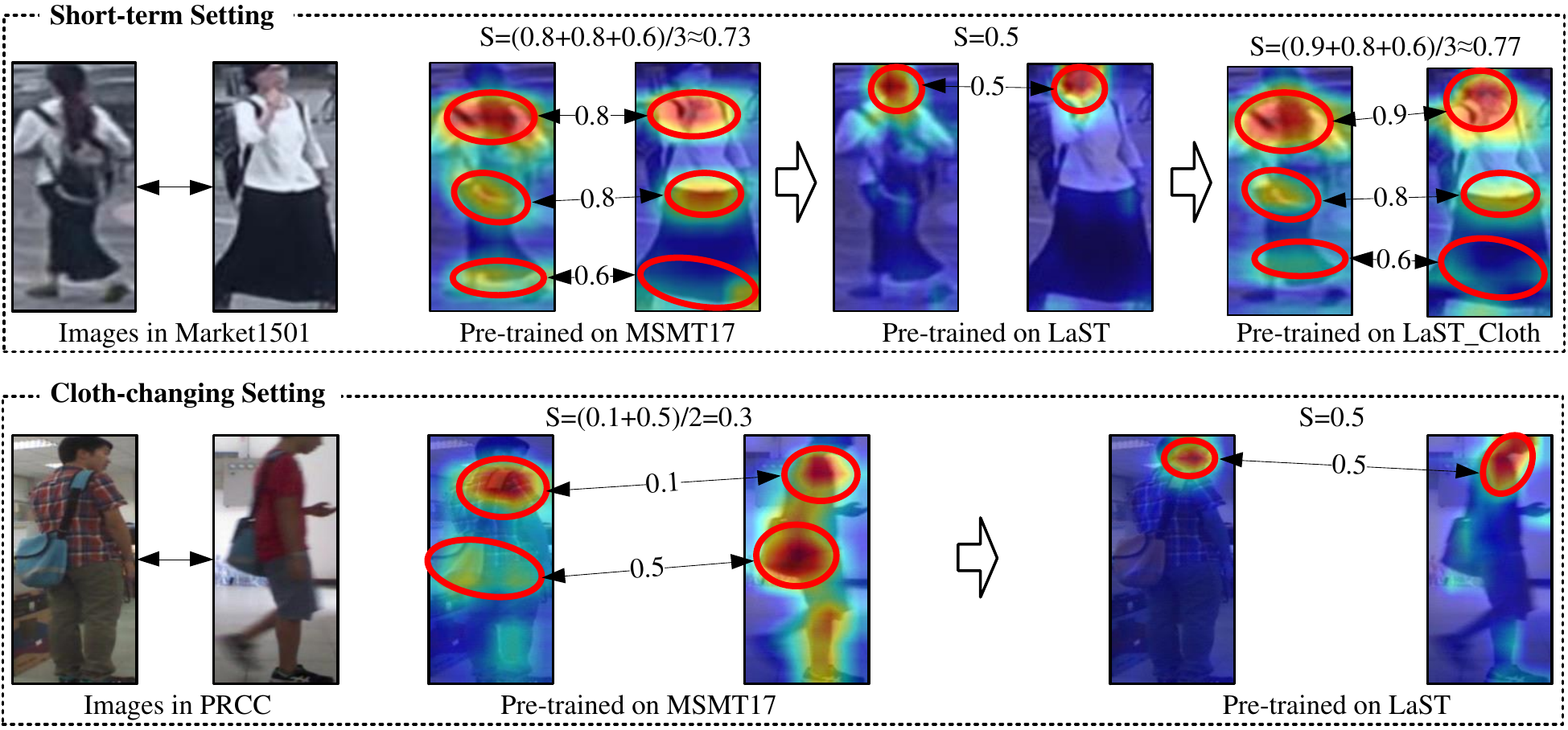} 
	\caption{\textbf{Similarity matching of learned parts among pre-trained models.} `S' denotes the similarity value. As more clothing cues can be used, the similarity values of pre-trained on MSMT17 and LaST\_Cloth are larger than that of pre-trained on LaST in the short-term setting. However, the learned clothing cues would deteriorate the performance in the cloth-changing setting. Therefore, the similarity value of pre-trained on MSMT17 is smaller than that of pre-trained on LaST.}	
	\label{fig:heat_analyse}
\end{figure*}

\subsection{Generalization of LaST on Short-term Setting}
\subsubsection{Performance Comparison}
Although LaST is designed for large-scale spatio-temporal person re-ID, we are still interested in its generalization ability in the short-term setting. Since MSMT17 is much larger than Market1501 and DukeMTMC, it is utilized as the pre-training dataset. For example, the model is first initialized with the parameters pre-trained on MSMT17, then it is trained and tested on Market1501 and DukeMTMC.

Table~\ref{table:transfer_short_term} shows that the performance of LaST is better than ImageNet but still poorer than MSMT17. It is because that the task of large-scale spatio-temporal person re-ID is different from conventional short-term person re-ID task. \textbf{In the short-term setting, pedestrians would not change their clothes and clothing cues are enough discriminative. However, clothing cues may not be reliable in the large-scale spatio-temporal settings}. The model trained with LaST mainly focuses on cloth-irrelevant cues. The lack of clothing features leads to the loss of generalization performance on short-term re-ID datasets. Therefore, the performance of LaST is inferior to MSMT17.

\subsubsection{Cloth-based Pre-training Strategy}
Based on the above analysis, we proposed a new pre-training strategy for the short-term setting. Each person in LaST may have several clothes. As shown in Fig.~\ref{fig:cloth_id}, the samples with different clothes are regarded as different labels, even if they belong to the same person. \textbf{The traditional training methods use pedestrian ID as the label, but the new pre-training strategy utilizes clothes as the label}. We denote the new pre-training strategy as ``LaST\_Cloth''. As shown in Table~\ref{table:transfer_short_term}, it significantly boosts the generalization performance and achieves the Rank1 accuracy of 93.1\% and 84.5\% on Market1501 and DukeMTMC, respectively. These experiments demonstrate the effectiveness of the new pre-training strategy. Therefore, we could utilize pedestrian labels for the cloth-changing setting and clothing labels for the short-term setting. This benefits from the dense annotation of clothes. LaST is the only person re-ID data set to label clothes at the present time.

\subsubsection{Evaluation on Identity Combination}
To further verify the effectiveness of the proposed cloth-based training strategy, we conducted the identity combination experiments in Table~\ref{table:transfer_comb}. For example, we combine MSMT17 and Market1501 as the training set. Since their identity numbers of training sets are 1,041 and 751, respectively, the total identity number of the combined dataset becomes 1,792. Then we train the combined dataset and evaluate the test set of Market1501. Table~\ref{table:transfer_comb} shows us that ``+LaST\_Cloth'' still achieves better performance in both Market1501 and DukeMTMC. This result further demonstrates the effectiveness of the cloth-based training strategy in short-term person re-identification.

\subsection{Visualization}
To verify the correctness of the above analysis, we visualize the heat maps in Fig.~\ref{fig:heat_map}. The class activation mapping (CAM) \cite{zhou2016learning} is used to generate the heat maps. The brighter the pixels are, the more attention the model pays to.

\subsubsection{Short-term Setting}
Three pre-trained models are utilized to visualize the heat maps on Market1501. As shown in Fig.~\ref{fig:heat_map}(a), the model trained with LaST mainly focuses on the heads and ignores the clothing cues. Since clothing cues are quite discriminative in the short-term setting, ignoring them would lead to poor performance. Benefitting from the cloth-based pre-training strategy, LaST\_Cloth could focus on the clothes in addition to the head cues. It further demonstrates the effectiveness of the proposed pre-training strategy. Fig.~\ref{fig:heat_map}(b) shows some instances when faces are invisible. It shows that the model trained with LaST could still focus on the head, especially the hair. 

We further give an illustration in Fig.~\ref{fig:heat_analyse}. In the short-term setting, the similarity of pre-trained on MSMT17 is much larger than that of pre-trained on LaST. This is because more clothing cues with high similarity can be exploited. The model pre-trained on LaST mainly focus on the head. As the area of head pixel is small, the similarity of the head is smaller than that of clothing cues. However, the model pre-trained on LaST\_Cloth could still learn clothing parts and achieve high match performance. The above visualizations fully explain why LaST behaves poorly in the short-term settings, while LaST\_Cloth and MSMT17 have better generalization performance in this scenario.

\subsubsection{Cloth-changing Setting}
Fig.~\ref{fig:heat_map}(c),(d) and (e) shows the visualization results on PRCC dataset. The pedestrians in cameras A and B have the same clothes but are different from the camera C. We could see that the models trained with Market1501 and MSMT17 still focus on the clothes even the clothes have been changed. This leads to their poor performance in cloth-changing settings. However, the model trained with LaST mainly focuses on the heads, which are more robust than clothes in this scenario. As shown in Fig.~\ref{fig:heat_analyse}, the similarity values of clothing cues are small because the boy has changed his clothes. As the clothing cues are not still reliable, the learned clothing cues would deteriorate the performance. The similarity of pre-trained on MSMT is smaller than that of pre-trained on LaST. Therefore, LaST has a better generalization ability than Market1501 and DukeMTMC in cloth-changing settings.

\subsection{Limitation and Discussion}
The performance of person re-ID on short-term datasets is satisfied, but it has not been widely used in real applications to date. A partial reason is its not robust performance in the large-scale spatio-temporal scenes. This task has not been well studied because of a lack of datasets. In this work, we provide LaST to do research in this scenario. Pedestrians in LaST have large spatio-temporal spans. This setting provides many challenging problems. For example, some pedestrians appear both daytime and night, and some frequently change their clothes. 
Besides, LaST provides sufficient diversity in terms of pedestrians, clothes, age, scenes, and weather, {\itshape etc}. The experiments show us that most current methods achieve unsatisfied performance on LaST, especially the poor mAP value. To solve the problem, more discriminative cues need to be studied. For example, the combination of body shape and discriminative appearance cues. It needs to be explored in the future.

\section{Conclusions}

This work studies large-scale spatio-temporal person re-identification. This task has much larger spatial and temporal spans than previous settings. Our major contribution is the large-scale benchmark dataset called LaST. It is the largest densely annotated re-ID benchmark and the first one to label clothes to date. By careful collection, the style of LaST is very similar to conventional re-ID datasets. Besides, we propose an simple but effective baseline that works well on such challenging person re-ID setting. Specifically, the mAP is directly optimized  during training and achieves competitive performance compared with current methods. By conducting extensive experiments, we demonstrate that LaST has good generalization ability in both short-term and cloth-changing scenarios. We believe that there is still much room for improvement in the large-scale spatio-temporal settings. By releasing LaST, we expect this dataset to catalyze research in the re-ID community and propel the maturation of re-ID techniques in real-world applications.



%
%

\ifCLASSOPTIONcaptionsoff
  \newpage
\fi



\bibliographystyle{IEEEtran}
\bibliography{ref_conf}

\end{document}